\documentclass[journal]{IEEEtran}

\usepackage{caption}
\usepackage{subcaption}
\ifCLASSINFOpdf
  \usepackage[pdftex]{graphicx}
  \graphicspath{{./graph/}}
  \DeclareGraphicsExtensions{.pdf,.jpeg,.png}
\else
  \usepackage[dvips]{graphicx}
  \graphicspath{{./graph/}}
  \DeclareGraphicsExtensions{.eps}
\fi

\ifCLASSINFOpdf
\else
\fi

\usepackage[cmex10]{amsmath,mathtools}
\usepackage{mdwmath}
\usepackage{mdwtab}
\usepackage{amssymb}
\usepackage{booktabs}
\usepackage{multirow}

\usepackage{tabularx}
\usepackage{algorithmic}
\usepackage{algorithm}
\usepackage{float}

\newcommand{\algorithmicinput}{\textbf{Input:}}
\newcommand\INPUT{\item[\algorithmicinput]}

\usepackage{epstopdf}

\usepackage{color}

\usepackage[utf8]{inputenc}

\usepackage{cite}

\usepackage{url}
\usepackage{hyperref}

\begin{document}

\title{Byzantine-Resilient Federated Large Language Model via Hypergraph Message Passing}

\author{
Kai~Li, \IEEEmembership{Senior Member, IEEE},
Hanlin Cai, \IEEEmembership{Member, IEEE},
Zihao Liu,
Yichen Li, \IEEEmembership{Member, IEEE}, \\
Jose Alcaraz Calero, \IEEEmembership{Senior Member, IEEE},
Wei Ni, \IEEEmembership{Fellow, IEEE}, 
Falko Dressler, \IEEEmembership{Fellow, IEEE},
and \\
Özgür B. Akan, \IEEEmembership{Fellow, IEEE}

\thanks{
K.~Li and J.~Calero are with the School of Computer Science and Digital Technologies, Aston University, Birmingham, UK (E-mail: kaili@ieee.org \& j.alcarazcalero@aston.ac.uk).

H.~Cai, Z.~Liu, Y.~Li, and O.~B.~Akan are with Centre for neXt Communications (CXC), Department of Engineering, University of Cambridge, CB3 0FA Cambridge, U.K., and O.~B.~Akan is also with the Centre for neXt Communications (CXC), Department of Electrical and Electronics Engineering, Ko\c{c} University, 34450 Istanbul, T\"{u}rkiye (e-mail: \{hc663, zl600, kl596, oba21\}@cam.ac.uk \& akan@ku.edu.tr).

W.~Ni is with the School of Engineering, Edith Cowan University, Perth, WA 6027, and the School of Computer Science and Engineering, University of New South Wales, Sydney, NSW 2052, Australia (E-mail: wei.ni@ieee.org).

F.~Dressler is with the Telecommunication Networks Group (TKN), School of Electrical Engineering and Computer Science, TU Berlin, Germany (E-mail: dressler@ccs-labs.org).
}

\thanks{\textbf{K.~Li and H.~Cai contributed equally to this work.}}

}


\IEEEcompsoctitleabstractindextext{%
\begin{abstract}
Federated Large Language Models (FedLLMs) are promising to deliver privacy-preserving, collaborative intelligence. The decentralized training process is vulnerable to Byzantine attacks that corrupt semantic consistency and factual resilience through malicious model updates. This paper presents a new resilience strategy that prevents semantically corrupted knowledge from propagating during federated aggregation. A defense problem is formulated to jointly maximize global accuracy and suppress manipulated model updates, enabling trustworthy knowledge aggregation without accessing private local data. To strengthen the resilience of FedLLMs, we develop a novel Hypergraph Message Passing Graph Autoencoder (HMP-GAE) framework, where low-rank model updates are represented as nodes in a hypergraph and high-order semantic relationships among multiple agents are exploited through iterative message passing. By leveraging historical checkpoints and cross-agent relational reasoning, HMP-GAE identifies and suppresses semantically inconsistent updates while preserving knowledge alignment among agents throughout federated training. HMP-GAE is implemented using Qwen-2.5 and Llama-3.2 as pretrained FedLLM backbones and evaluated on AG News and Yahoo! Answers datasets. Compared to existing defenses, HMP-GAE improves global accuracy by $1.21$\%, reduces final-model classification semantic entropy by $44.08$\%, and lowers perplexity by $13.34$\%. HMP-GAE outperforms the GAE-based defense by improving the accuracy $1.39$\%, reducing the semantic entropy by $54.5$\%, and lowering perplexity by $35.5$\%.
\end{abstract}

\begin{keywords}
Byzantine Attacks, Federated Large Language Models (FedLLMs), Graph Autoencoder (GAE), Hypergraph Message Passing, CyberEdge
\end{keywords}}

\maketitle

\IEEEdisplaynotcompsoctitleabstractindextext
\IEEEpeerreviewmaketitle

\section{Introduction}
Large language models (LLMs) are emerging as a transformative paradigm for intelligent, autonomous operations in CyberEdge networks~\cite{cai2025graph,li2025future}, where distributed devices, from IoT sensors to connected vehicles and industrial robots, collaborate securely and efficiently without centralized control.
By integrating federated learning (FL) with LLMs, edge nodes can perform local reasoning, decision-making, and adaptation based on their contextual data while sharing high-level knowledge or model updates rather than raw information~\cite{qiao2025survey,cheng2024towards,li2025towards}. This enables a new generation of self-organizing, privacy-preserving, and resilient CyberEdge networks that can support complex, real-time applications, e.g., autonomous mobility, digital twin synchronization, smart healthcare, and disaster response.
Federated LLMs (FedLLMs) act as cognitive coordinators, learning human-aligned behaviors, negotiating across agents, and dynamically optimizing communication, computation, and sensing processes under constrained resources~\cite{yao2024federated,webb2025brain,yang2025llm,cai2026graph1}. This synergy enhances situational awareness, interoperability, and trust across heterogeneous edge environments, paving the way to autonomous, explainable, and resilient artificial intelligence (AI) infrastructures at the network edge.

In FedLLMs, each edge device or agent maintains a local instance of an LLM that learns from its own private data, such as text or sensory streams, arising from its specific environment and task context. Instead of transmitting raw data to a central server, participants share only model updates or gradients through secure aggregation, collaboratively refining a global foundation model. The nature of FedLLMs empowers each agent to integrate and reason across diverse data features, dynamically adapt its learning strategy to evolving local contexts, and coordinate with other agents based on shared semantics, trust levels, and task interdependencies~\cite{cai2024securing, wang2025federated,hu2024federated,cai2026graph2}. Through iterative aggregation and local fine-tuning, FedLLMs collectively evolve into a distributed intelligence capable of robust language understanding, cross-modal reasoning, and context-aware decision-making in heterogeneous and privacy-sensitive CyberEdge environments.

Within this collaborative framework, FedLLMs are particularly vulnerable to Byzantine attacks, where adversaries exploit the open-ended generative capacity of LLMs and their internal representation spaces to propagate false or biased knowledge across agents~\cite{ren2025advances,uddin2025systematic, cai2026poster}.

\begin{figure}
    \centering
    \includegraphics[width=1\columnwidth]{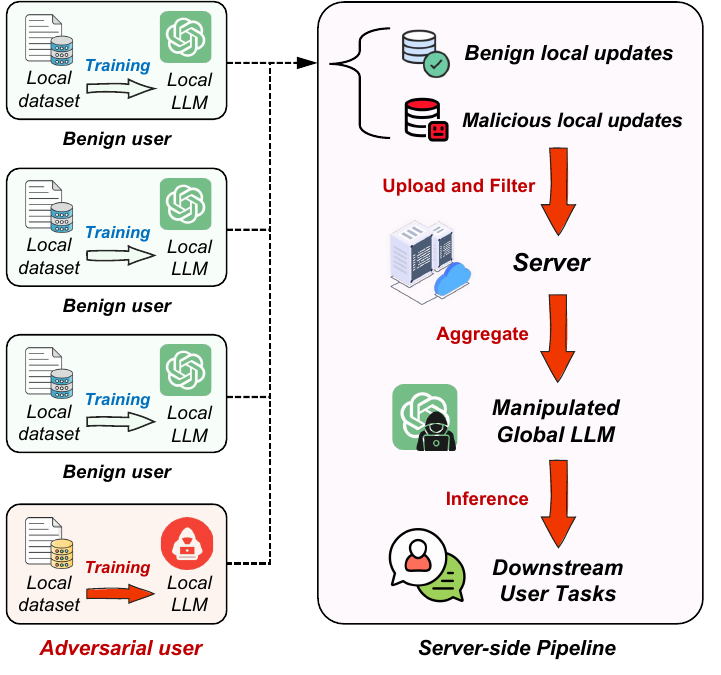}
    \caption{\small In FedLLMs, users train their local LLMs and produce local model updates.}
\label{fig_ByzantineAttacks}
\end{figure}

As depicted in Fig.~\ref{fig_ByzantineAttacks}, a malicious client or compromised edge agent may inject subtle cross-modal perturbations, such as corrupted textual prompts, mislabeled visual samples, or biased sensory data, into its local training process, steering the local model toward distorted semantic associations or fabricated knowledge~\cite{huang2024federated}. When these adversarial updates are aggregated, the global model's latent graph and contextual alignments be corrupted, leading to systemic manipulations that produce confident yet false textual descriptions, visual interpretations, or sensor-based inferences. Over multiple communication rounds, these manipulations can cascade through the network, undermining the collective intelligence of FedLLMs. For instance, a malicious agent in a vehicular CyberEdge network might cause an LLM to create nonexistent obstacles from camera feeds, while in healthcare or industrial IoT systems, fabricated insights could misguide diagnostics or trigger unsafe automated actions.

Moreover, the distributed and privacy-preserving nature of FedLLMs amplifies the impact of Byzantine attacks, since each agent trains locally on private data, and centralized verification or direct inspection of training samples is infeasible. A single compromised agent can influence others through conversational interactions, aggregation of model updates, or semantic alignment mechanisms, gradually inducing large-scale semantic drift in the collective model. Such drift erodes the interpretability, factual consistency, and ethical resilience of the FedLLMs.

This paper presents a new Byzantine-resilient strategy to mitigate Byzantine attacks in FedLLMs that distort semantic consistency and induce factual misalignment in LLMs. The strategy introduces a self-healing mechanism within the federated aggregation process to recognize and neutralize adversarial model updates before they propagate semantic corruption across FedLLMs. A problem is cast to maximize task accuracy of the LLMs while simultaneously identifying and suppressing manipulated updates. This filters for trustworthy and semantically coherent knowledge, thereby improving the learned representations.

To adaptively enhance the resilience of FedLLMs against latent Byzantine-inducing perturbations, we design the \textit{Hypergraph Message Passing Graph Autoencoder (HMP-GAE) framework}. In HMP-GAE, at the server each agent's model update that is fine-tuned by low-rank adaptation (LoRA) is represented as a node within a hypergraph, where hyperedges capture high-order semantic and contextual relationships among multiple agent updates, as opposed to simple pairwise correlations. The server performs iterative message passing across these hyperedges to refine the embeddings of benign updates, allowing collective reasoning that effectively identifies and suppresses corrupted or semantically inconsistent contributions. By leveraging hypergraph structures, HMP-GAE distills reliable relational knowledge from trustworthy agents and historical checkpoints, ensuring robust aggregation that maintains factual coherence and mitigates adversarial manipulation.

The contributions of this paper are listed as follows: 
\begin{itemize}
    \item This paper proposes the first graph neural network-based strategy for mitigating Byzantine attacks in FedLLMs, which enhances the resilience of FedLLMs against subtle adversarial perturbations. The strategy introduces a defense perspective that focuses on preserving semantic integrity and factual consistency across distributed FL. We formulate a defense problem at benign users to maximize global task accuracy while filtering out malicious, semantically corrupted contributions during aggregation.

    \item We design the HMP-GAE framework to enhance FedLLMs against Byzantine attacks through structured relational reasoning and representation alignment. By modeling each agent's update as a node in a hypergraph and capturing high-order semantic relationships via message passing, HMP-GAE identifies and suppresses semantically inconsistent contributions. This hypergraph-based aggregation mechanism strengthens cross-agent knowledge coherence, preserves reasoning integrity, and mitigates adversarial manipulation.

    \item Extensive experiments are conducted under diverse adversarial scenarios to evaluate the HMP-GAE framework. We assess key metrics, including prediction accuracy, semantic entropy, and perplexity. HMP-GAE achieves superior accuracy and resilience while maintaining low resource overhead, validating its practicality as an effective and scalable Byzantine-resilient defense for FedLLMs under malicious model updates. The source code of the proposed HMP-GAE is released on GitHub: \href{https://github.com/GuangLun2000/HMP-GAE}{https://github.com/GuangLun2000/HMP-GAE}.
\end{itemize}

The remainder of this paper is structured as follows. Section~\ref{sec_review} reviews the literature on Byzantine-resilient mitigation, assessment, and detection in FL and LLMs. Section~\ref{sec_FALLMs} investigates the state-of-the-art training architecture of FedLLMs as well as the threat model of Byzantine attacks. The proposed HMP-GAE for mitigating Byzantine attacks is described in Section~\ref{sec_HMP-GAE}. Section~\ref{sec_performance} presents the experimental implementation in PyTorch, and resilience performance analysis.  Section~\ref{sec_cond} concludes this paper.

\section{Related Work}
\label{sec_review}
This section reviews the existing studies on Byzantine attacks mitigation and detection in FL. Our contributions over the existing Byzantine-resilient models are presented.

\subsection{Byzantine-resilient Strategies}

The authors in~\cite{xia2026communication} developed a communication-efficient and Byzantine-tolerant FL in CyberEdge networks. It combined sign-refined momentum updates with error compensation to reduce communication costs and stabilize convergence. Edge servers used cosine similarity to assess trust and suppress unreliable local updates. The cloud server combined real and dummy data projections through projection consensus for global aggregation. 
Layer-wise cosine similarity was used in~\cite{wang2026byzantine} to detect malicious gradients against a gradient median benchmark. A proactive defense was formulated to constrain divergence among benign gradients, while trust scores reduce misjudgments under non-IID data. 

In~\cite{li2026model}, a Byzantine-resilient FL was studied for improving global model consistency. An asynchronous common subset protocol was presented, where benign users can aggregate the model updates despite conflicting messages from malicious users. Dual-domain trust scoring was used to evaluate consensus behavior, similarity among model updates, and stability across rounds to limit malicious influence. Shen et al.~\cite{shen2026byzantine} presented a Byzantine-resilient FL, which can distinguish malicious model updates through discrepancies in local models' feature representations. A reference model trained on public data was utilized to quantify these discrepancies as feature fingerprints. These fingerprints were clustered to filter out malicious models before aggregation. 

Two FL schemes were developed in~\cite{xia2026perfect} to enhance resilience against Byzantine users while preventing eavesdroppers from learning users' private data. Both used a small representative dataset at the server and a discriminator function to suppress malicious contributions. Lagrange coded computing was combined with re-randomization to protect privacy, while communication costs were reduced through a trusted third party. Moreover, a sparsity-guided refinement scheme could be added to mitigate Byzantine attacks~\cite{zhang2026building}. Sparsity was utilized to derive significance distribution vectors that expose subtle malicious changes in client model updates. These vectors are clustered to compute weighted similarity scores for group resilience. The resulting groups guide further refinement of raw updates before robust aggregation. 

Bray-Curtis dissimilarity was introduced in~\cite{zhu2026bpflh} to measure element-wise differences among users' gradients. Their method can reduce the misclassification of benign gradients against Byzantine attacks under non-IID data distributions. A lattice-based fully homomorphic encryption scheme was used to protect local gradients while enabling secure aggregation and Byzantine user detection. 
The authors in~\cite{qin2026efficient} formulated a Byzantine-resilient FL scheme combining homomorphic encryption with gradient dimension compression. The Johnson-Lindenstrauss transformation was applied to reduce gradient dimensions while preserving geometric relationships needed for Byzantine defense. 

Existing Byzantine-resilient aggregation rules were adapted in~\cite{tang2025roby}. A shared, updated consensus dataset was introduced to provide a common benchmark for applying these rules and reaching agreement. A dual-layer privacy shielding strategy was designed to protect local models while preserving global model accuracy. 
A Byzantine-resilient FL framework was developed in~\cite{song2025byzantine}, combining client-side and server-side defenses. Clients used a model-level penalty loss to limit divergence from benign models and reduce correlation with malicious model updates, which can help prevent benign users from being misclassified. The server combined manifold-based analysis with a maximum-clique strategy to strengthen Byzantine detection.

Wang et al.~\cite{wang2025pdfl} presented a Byzantine-resilient scheme to defend FL against poisoning attacks while protecting user privacy. A two-server architecture used secure two-party computation to analyze local model gradients.  A security protocol was designed to normalize users' model gradients, compute their cosine similarities, and identify malicious model updates. 
In~\cite{xiao2025flgym}, a two-stage defense framework was presented for Byzantine-resilient FL. Malicious users can be detected through model similarity. A validation mechanism was then used to estimate similarity among users' local data to refine detection decisions. A weight recovery mechanism was also applied to adjust the aggregation weights of users identified as Byzantine.

In~\cite{zhang2025practical}, dimensionality reduction was adopted to accelerate the private computation of complex filtering rules in Byzantine-resilient FL. An adaptive tuning strategy was formulated to limit the influence of malicious model updates that evade filtering in the reduced space. 
Moreover, the authors in~\cite{yu2025lpp} introduced a correction term during local training to reduce model inconsistency. A secure protocol deployed across two servers enabled Byzantine-resilient aggregation while protecting user privacy, which was expected to limit the privacy computation burden on resource-constrained users. 

Herath et al.~\cite{herath2025dsfl} developed a dual-server protocol to protect local model updates without encryption or key exchange. A group-wise filtering mechanism used deviation scores and client credits to isolate Byzantine users, where a reward-penalty system can promote fair participation throughout training. 
To address heavy-tailed gradient noise, the authors in~\cite{tao2026byzantine} combined distributed stochastic heavy-ball optimization with update normalization and gradient clipping. Nearest-neighbor mixing through random projections was used to isolate the impact of Byzantine users.

An internal auditor was built in~\cite{yazdinejad2024robust} to examine gradient similarity and distribution to distinguish benign model updates from malicious ones. It combined a Gaussian mixture model with Mahalanobis distance for Byzantine-tolerant aggregation. 
A three-server computation was developed to perform Byzantine-resilient aggregation~\cite{dong2023privacy}, where each server held two of the three shares for privacy protection. A top-k protocol was invoked to measure the cosine distance and compute the scores to select benign local models.

\subsection{Our Contributions}
Existing Byzantine-resilient techniques primarily address Byzantine attacks in FL. Their defenses employ robust aggregation, model update similarity, trust evaluation, and secure computation to identify malicious contributions. However, Byzantine resilience in FedLLMs has not been specifically addressed. Our work aims to investigate Byzantine-resilient FedLLMs, focusing on how malicious updates disrupt semantic consistency and knowledge alignment during federated aggregation.

Moreover, our proposed HMP-GAE framework for Byzantine-resilient FedLLMs enables each user's LoRA update to form a hypergraph node, while hyperedges capture semantic relationships among multiple users. HMP-GAE combines historical checkpoints with iterative hypergraph message passing to learn shared latent representations of these updates. HMP-GAE captures high-order relational inconsistencies that individual model update analysis may overlook. Therefore, the representations are leveraged to identify and suppress semantically inconsistent updates during aggregation, which enables Byzantine-resilient knowledge integration without accessing private local data.

\section{Formulation of FedLLMs and Threat Model}
\label{sec_FALLMs}
In this section, we first formulate collaborative adaptation of a shared pretrained LLM and then describe Byzantine manipulation of the aggregated
parameters.

\subsection{Fine-tuning of FedLLMs}

Consider FedLLMs with $K\geq2$ agents participating in each of $R$ communication rounds, where $r=0,1,\ldots,R-1$. Let $\theta\in\mathbb{R}^{d_\theta}$ denote the frozen pretrained backbone parameters that are not updated during federated fine-tuning. The benign and Byzantine agents form disjoint sets $\mathcal{B}$ and $\mathcal{A}$, respectively, with $\mathcal{B}\cup\mathcal{A}=\{1,\ldots,K\}$. Agent $k$ holds a private dataset $\{(x_{k,n},y_{k,n})\}_{n=1}^{N_k}$, where $N_k>0$ is the number of data samples at the agent, $x_{k,n}$ is the input text, and $y_{k,n}=(y_{k,n}^{1},\ldots,y_{k,n}^{T_{k,n}})$ is the target token sequence of length $T_{k,n}$. 

To condition the LLM, the input encoder $f_\phi$ and the context projection module $g_\psi$ produce~\cite{he2024g}
\begin{align}
c_{k,n}=g_\psi(f_\phi(x_{k,n})),
\end{align}
where the trainable parameters $\phi$ and $\psi$ are separate from the frozen backbone $\theta$. 

We adopt LoRA for the backbone. Let $\mathcal{S}$ index the adapted projections. For each $\ell \in \mathcal{S}$, the pre-trained matrix $W_\ell\in\mathbb{R}^{a_\ell\times b_\ell}$ remains frozen, and its update $\Delta W_\ell$ is parameterized as
\begin{align}
\Delta W_\ell&=\frac{\alpha_\ell}{q_\ell}B_\ell A_\ell, \label{eq_lora_parameterization}\\
A_\ell&\in\mathbb{R}^{q_\ell\times b_\ell}, \;\;\; B_\ell\in\mathbb{R}^{a_\ell\times q_\ell}.
\nonumber
\end{align}
where $a_\ell$ and $b_\ell$ denote the output and input
dimensions of projection $\ell$, respectively; $q_\ell\ll\min(a_\ell,b_\ell)$ is the adaptation rank; and $\alpha_\ell$ is a scaling parameter. The trainable factors $A_\ell$ and $B_\ell$ map the input to a rank-$q_\ell$ space and then to the output space, respectively. 

Let $\lambda=\{A_\ell,B_\ell\}_{\ell\in\mathcal{S}}$ collect the LoRA factors. The backbone update $\Delta\theta(\lambda)$ places the vectorized matrices $\Delta W_\ell$ at their corresponding parameter locations and is zero elsewhere. Thus, the backbone is $\theta+\Delta\theta(\lambda)$, and the trainable parameter collection is given by
\begin{align}
\omega=(\lambda,\phi,\psi),
\label{eq_omega}
\end{align}

The model's predictive distribution can be given as $p_{\theta,\omega}$, and autoregressive factorization is
\begin{align}
p_{\theta,\omega}(y_{k,n}\!\mid \!x_{k,n})\! &=\!\prod_{j=1}^{T_{k,n}} p(y_{k,n}^{j}\!\mid \!y_{k,n}^{<j},c_{k,n}; \theta\!+\!\Delta\theta(\lambda)),
\label{eq_autoregressive}
\end{align}
where $j$ indexes target tokens, and $y_{k,n}^{<j}$ contains the preceding tokens. The local objective averages sequence negative log-likelihoods across samples following
\begin{align}
\mathcal{L}_k(\omega) =-\frac{1}{N_k}\sum_{n=1}^{N_k}\sum_{j=1}^{T_{k,n}} \log p(y_{k,n}^{j}\mid y_{k,n}^{<j},c_{k,n}; \theta+\Delta\theta(\lambda)).
\label{eq_local_loss}
\end{align}
Thus, the FedLLMs' objective can be formulated as
\begin{align}
\min_\omega F(\omega)&=\sum_{k=1}^{K}p_k\mathcal{L}_k(\omega) \;\text{ with }\;\label{eq_global_objective}
p_k=\frac{N_k}{\sum_{i=1}^{K}N_i}, 
\end{align}
where $p_k$ describes the intended collaborative objective.

At communication round $r$, the server broadcasts $\omega^r$. Each benign agent $k\in \mathcal{B}$ initializes $\omega_k^{r,0}=\omega^r$ and performs $E$ local stochastic gradient descent steps as
\begin{align}
\omega_k^{r,e+1} &=\omega_k^{r,e}-\eta_{r,e} \nabla_\omega\widehat{\mathcal{L}}_k (\omega_k^{r,e};\mathcal{M}_k^{r,e}), 
\label{eq_gradientdescent_lora}\\
e&=0,1,\ldots,E-1,\nonumber
\end{align}
where $\eta_{r,e}$ is the learning rate, $\mathcal{M}_k^{r,e}$ is a sampled minibatch, and $\widehat{\mathcal{L}}_k$ is its loss estimate.
The model update from agent $k$ is denoted by
\begin{align}
\widehat{\omega}_k^r =(\{\widehat{A}_{k,\ell}^r,\widehat{B}_{k,\ell}^r\}_{\ell\in\mathcal{S}}, \widehat{\phi}_k^r,\widehat{\psi}_k^r).
\end{align}
We have $\widehat{\omega}_k^r=\omega_k^{r,E}$, $\forall k\in\mathcal{B}$.

Given $\ell \in \mathcal S$, the server assigns weights $\tilde p_k^r\geq0$ that have to satisfy $\sum_{k=1}^{K}\tilde p_k^r=1$. The additional trainable modules adopt the same weights
\begin{align}
\phi^{r+1}&=\sum_{k=1}^{K}\tilde p_k^r\widehat\phi_k^r, \;\;\; \psi^{r+1}=\sum_{k=1}^{K}\tilde p_k^r\widehat\psi_k^r.
\label{eq_aggregate_modules}
\end{align}
Therefore, the global backbone is given as
\begin{align}
\theta^{r+1} =\theta+\Delta\theta(\lambda^{r+1}).
\label{eq_effective_backbone}
\end{align}

\subsection{Threat Model of Byzantine Attacks}

Compromised agents, i.e., agent $a \in \mathcal{A}$, can upload arbitrary finite parameters, subject to the proximity constraint specified below. They receive the broadcast global models but cannot access the benign agents' private datasets. Let $\mathcal Q$ denote an evaluation distribution over input-target pairs $(x,y)$. The averages attack success rate over $\mathcal Q$ can be defined as
\begin{align}
\mathcal R^{\mathrm{adv}}(\omega) &= \mathbb E_{(x,y)\sim\mathcal Q} [h_{\mathrm{adv}}(p_{\theta,\omega}(\cdot\mid x),y)],
\label{eq_adversarial_risk}
\end{align}

To reduce detectability, each malicious agent $a \in \mathcal{A}$ constrains the deviation of its LoRA weights from those of the current broadcast global model. At communication round $r$, let $\widehat B_{a,\ell}^r$ and $\widehat A_{a,\ell}^r$, $\forall \ell \in \mathcal{S}$ denote the LoRA factors of malicious agent $a\in\mathcal{A}$. The matrices $B_\ell^r$ and $A_\ell^r$ are the global LoRA factors parameterizing the weight update to the frozen pretrained layer $\ell$ at round $r$. According to~\cite{yan2026fedsrd,wang2024flora}, the weight change at adapted layer $\ell$ is given by 
\begin{align}
U_{a,\ell}^r &= \frac{\alpha_\ell}{q_\ell} (\widehat B_{a,\ell}^r\widehat A_{a,\ell}^r - B_\ell^r A_\ell^r).
\label{eq_malicious_change}
\end{align}

At communication round $r$, the malicious agents seek to maximize the risk of the resulting aggregated model as
\begin{align}
&\max_{\{\widehat\omega_a^r\}_{a\in\mathcal A}}
\mathcal R^{\mathrm{adv}}(\omega^{r+1})
\nonumber\\
&\text{s.t.} \;\;\; \omega^{r+1} = \sum_{k=1}^{K}\tilde p_k^r\widehat\omega_k^r, \\
&\qquad \widehat\omega_k^r=\omega_k^{r,E}, \\
&\qquad \sum_{\ell}\|U_{a,\ell}^r\|_F^2 \leq \varepsilon^2, \;\forall a\in\mathcal A. \label{eq_optimalThreat}
\end{align}
Constraint~\eqref{eq_optimalThreat} runs over LoRA-adapted layers. $\|\cdot\|_F$ denotes the Frobenius norm. $\varepsilon>0$ bounds the total deviation in the LoRA weights. 

\begin{figure*}[t]
\centering
\includegraphics[width=1.72\columnwidth]{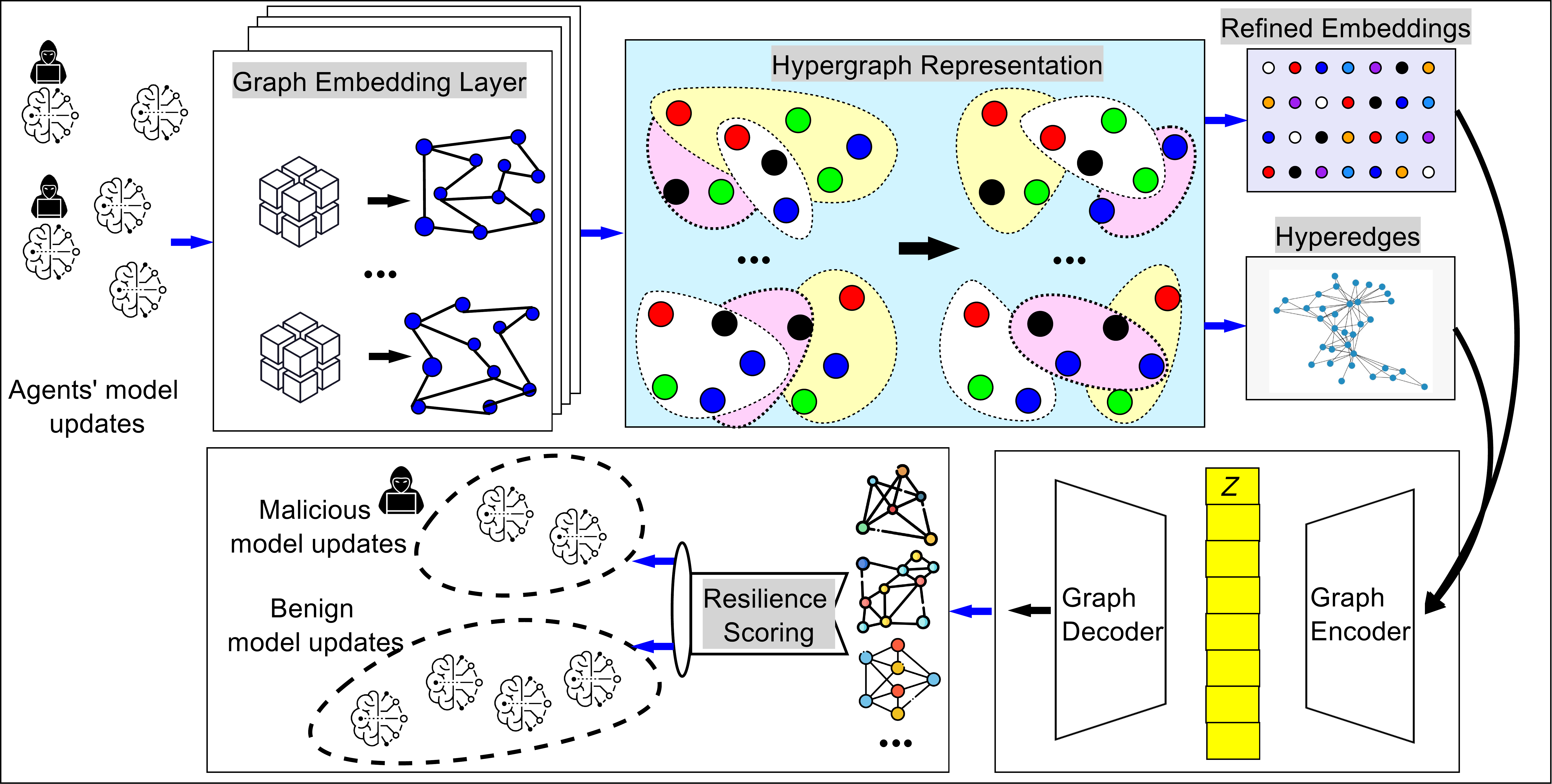}
\caption{\small The architecture of the proposed HMP-GAE framework, which is applied at the server. By modeling each agent's update as a node in a hypergraph and capturing high-order semantic relationships via message passing, HMP is developed with GAE to identify and suppress semantically inconsistent or malicious contributions.}
\label{fig_Framework}
\end{figure*}

\section{HMP-GAE for Byzantine-Resilient FedLLMs}
\label{sec_HMP-GAE}

In this section, HMP-GAE is designed to assess aggregated model updates before aggregation, which combines HMP with GAE to capture relationships among multiple agents and identify potentially inconsistent contributions. The server uses aggregated parameters and historical checkpoints without accessing private local training data.

\subsection{Architecture of the Proposed HMP-GAE}

As shown in Fig.~\ref{fig_Framework}, the server processes $K$ model updates through the graph embedding layer, where the benign and Byzantine identities are unknown to the defense. Let $\Theta$ collect the trainable parameters of HMP-GAE, including its feature embedding network, GAE encoders, and decoders. 

\subsubsection{Update Representation}

The server can reconstruct the change in weights relative to the broadcast global model
\begin{align}
U_{k,\ell}^r=\frac{\alpha_\ell}{q_\ell} (\widehat B_{k,\ell}^r\widehat A_{k,\ell}^r-B_\ell^r A_\ell^r),
\label{eq_reconstructed_change}
\end{align}
where $\widehat A_{k,\ell}^r$ and $\widehat B_{k,\ell}^r$ are the LoRA factor matrices uploaded by agent $k$.

Furthermore, each reconstructed matrix $U_{k,\ell}^r$ measures the difference between the additive weight updates represented by agent $k$'s LoRA factors and the global LoRA factors. The layer-wise differences can be concatenated with the changes in the input encoder parameters $\phi$ and context projection parameters $\psi$ to form a single descriptor of the agent's model update.
Let $\operatorname{vec}(\cdot)$ denote column-wise vectorization, and the LLM descriptor can be formulated as
\begin{align}
u_k^r=\operatorname{concat}\big( &\{\operatorname{vec}(U_{k,\ell}^r)\}_{\ell\in\mathcal S}, \nonumber\\
&\operatorname{vec}(\widehat\phi_k^r-\phi^r), \operatorname{vec}(\widehat\psi_k^r-\psi^r)\big) \in\mathbb R^D,
\label{eq_update_descriptor}
\end{align}
where $D$ represents the descriptor's dimension. 

For a uniform architectural description, all node and hyperedge embeddings have width $d$. A feature network as a trainable mapping can combine the model-update descriptor $u_k^r$ with the historical embedding $h_k^{r-1}\in\mathbb R^d$, $h_k^{-1}=\mathbf0$, and map them into the shared $d$-dimensional feature space. Thus, we have
\begin{align}
\Omega_k^r=f_{\mathrm{enc},\xi}(u_k^r,h_k^{r-1})\in\mathbb R^d, \;\; k=1,\ldots,K,
\label{eq_nodefeature}
\end{align}
where $\xi\subseteq\Theta$ denotes the parameters. 
Then, stacking the node features gives
\begin{align}
X^r=[\Omega_1^r,\ldots,\Omega_K^r]^\top\in\mathbb R^{K\times d}.
\label{eq_feature}
\end{align}

\subsubsection{Hypergraph Construction}
Since a hyperedge can jointly connect multiple agents with related update representations. This allows HMP-GAE to assess each update within a group of agents and capture relational patterns that isolated or pairwise comparisons may overlook. Moreover, an agent can belong to multiple overlapping hyperedges, allowing its update to be evaluated across different groups.

At communication round~$r$, the hypergraph $\mathcal H^r=(\mathcal V^r,\mathcal E^r)$ contains one node $v_k$ per agent. 

Let $\mathcal N_s^r(k)$ contain the $s$ nearest nodes to node $v_k$ under Euclidean distance or Cosine similarity between the initial node features, where $1\leq s < K$ and ties are resolved by agent index. Given $K$ hyperedges, each agent defines a hyperedge
\begin{align}
e_k^r=\{v_k\}\cup\{v_i:i\in\mathcal N_s^r(k)\}.
\end{align}
The incidence matrix is $H^r\in\{0,1\}^{K\times K}$ with the $(i,k)$-th element given by
\begin{align}
H_{ik}^r= 
\begin{cases}
1,&\text{if } v_i\in e_k^r;\\
0,&\text{otherwise}.
\end{cases}
\label{eq_incidence}
\end{align}

For simplification, the round superscript $^r$ is omitted within the following message-passing and reconstruction equations. The degree matrices are defined by~\cite{zhang2025hypergct}
\begin{align}
(D_{\mathcal V})_{ii}&=\sum_{m=1}^{K}H_{im}, \;\; (D_{\mathcal E})_{mm}=\sum_{i=1}^{K}H_{im},
\label{eq_degree}
\end{align}
with zero off-diagonal entries. $m$ indexes hyperedges.

\subsubsection{Hypergraph Message Passing}
HMP is designed to assess each agent's update using evidence from other agents connected through shared hyperedges. Node-to-hyperedge aggregation summarizes their collective update patterns, while hyperedge-to-node propagation incorporates this context into each agent's embedding. The decoder then reconstructs pairwise similarities and hyperedge incidence from these embeddings. This provide reconstruction errors that indicate departures from the modeled relational structure.

Let $\nu=0,1,\ldots,L-1$ represent HMP layers. HMP alternates between node-to-hyperedge aggregation and hyperedge-to-node propagation following~\cite{telyatnikov2023hypergraph}
\begin{align}
E^{(\nu)}&=\rho(D_{\mathcal E}^{-1}H^\top Z^{(\nu)}W_{\mathcal E}^{(\nu)}), \label{eq_node2edge}\\
Z^{(\nu+1)}&=\rho(D_{\mathcal V}^{-1}H E^{(\nu)}W_{\mathcal V}^{(\nu)}), \label{eq_Z}
\end{align}
where $Z^{(0)}=X$ and $E^{(\nu)}, Z^{(\nu+1)} \in \mathbb R^{K\times d}$. The hidden activation $\rho$ can be ReLU. The output is $Z=Z^{(L)}\in\mathbb R^{K\times d}$.

\subsubsection{Decoder Reconstruction}

Let $\mathbf A$ denote graph adjacency. Then, we have 
\begin{align}
\mathbf A_{ij}=
\begin{cases} 
1,&\text{if }i\neq j\text{ and }(HH^\top)_{ij}>0;\\
0,&\text{otherwise}.
\end{cases}
\label{eq_adjacency_target}
\end{align}
The decoder estimates their connectivity as
\begin{align}
\widehat{\mathbf A}_{ij}=\operatorname{sigmoid}(z_i^\top z_j), \;\; i\neq j.
\label{eq_adjacency_reconstruction}
\end{align}
Moreover, hyperedge membership can be reconstructed by
\begin{align}
\widehat H=\operatorname{sigmoid} (ZW_{\mathrm{dec}}+\mathbf1 b_{\mathrm{dec}}^\top) \in(0,1)^{K\times K},
\label{eq_incidence_reconstruction}
\end{align}
where $W_{\mathrm{dec}}\in\mathbb R^{d\times K}$, $b_{\mathrm{dec}}\in\mathbb R^K$, and $\mathbf1\in\mathbb R^K$. 

Define $\operatorname{BCE}(a,b)=-a\log b-(1-a)\log(1-b)$. The losses are given by 
\begin{align}
\mathcal L_H&=\frac{1}{K^2}\sum_{i=1}^{K}\sum_{m=1}^{K} \operatorname{BCE}(H_{im},\widehat H_{im}),
\label{eq_agent_reconstruction1}\\
\mathcal L_A&=\frac{1}{K(K-1)}\sum_{i\neq j} \operatorname{BCE}(\mathbf A_{ij},\widehat{\mathbf A}_{ij}).
\label{eq_agent_reconstruction2}
\end{align}
A smoothness regularizer can be defined for neighboring representations, which gives
\begin{align}
\mathcal L_{\mathrm{sm}} =\frac{\sum_{i\neq j}\mathbf A_{ij}\|z_i-z_j\|_2^2} {\max(1,\sum_{i\neq j}\mathbf A_{ij})}.
\end{align}
The training objective is given by 
\begin{align}
\mathcal L_{\mathrm{GAE}} =\lambda_H\mathcal L_H+\lambda_A\mathcal L_A +\lambda_{\mathrm{sm}}\mathcal L_{\mathrm{sm}},
\label{eq_reconstructionloss}
\end{align}

\subsection{Algorithm Design of Resilience Strategy}

The novelty of HMP-GAE lies in modeling Byzantine resilience in FedLLMs as a high-order relational inference problem over LoRA updates. Each agent's model update is represented as a hypergraph node, allowing the server to assess its consistency with groups of agents instead of relying solely on isolated or pairwise comparisons. HMP and reconstruction jointly capture these cross-agent relationships, while historical consistency provides a temporal reference for assessing suspicious deviations. The resilience estimates control federated aggregation, linking relational manipulation detection to the suppression of potentially corrupted model updates.

Algorithm~\ref{alg_hmpgae} implements this strategy by combining decoder reconstruction errors with historical consistency to determine each agent's aggregation weight.

\subsubsection{Historical Consistency and Training}

The server can keep the checkpoint $\mathcal C^r=(\{u_k^r,h_k^{r-1}\}_{k=1}^{K},H^r)$, as well as the final embeddings and resilience scores. For communication round $r>0$, it re-encodes the preceding checkpoint with the
current encoder parameters as 
\begin{align}
\bar z_k^r=\operatorname{sg}
([\operatorname{Enc}_\Theta(\mathcal C^{r-1})]_k).
\label{eq_historical_reference}
\end{align}
Here, $\operatorname{Enc}_\Theta(\cdot)$ includes feature embedding and HMP, $[\cdot]_k$ extracts a column-vector node representation, and $\operatorname{sg}(\cdot)$ stops gradients through the reference branch. 

Let $\epsilon>0$ be a numerical stability constant. The current and reference embeddings can be normalized as
\begin{align}
\widetilde z_k^r&=\frac{z_k^r}{\|z_k^r\|_2+\epsilon}, \;\; \widetilde{\bar z}_k^r =\frac{\bar z_k^r}{\|\bar z_k^r\|_2+\epsilon}.
\end{align}
Their historical discrepancy can be determined by $\chi_k^r=\|\widetilde z_k^r-\widetilde{\bar z}_k^r\|_2^2$. 
Based on the preceding resilience scores $\mathcal Z_k^{r-1}$, the historical regularizer can be given by
\begin{align}
\mathcal L_{\mathrm{hist}}^r =\frac{\sum_{k=1}^{K}\mathcal Z_k^{r-1}\chi_k^r} {\epsilon+\sum_{k=1}^{K}\mathcal Z_k^{r-1}}.
\label{eq_historical_loss}
\end{align}
At $r=0$, we set $\chi_k^0=0$ and $\mathcal L_{\mathrm{hist}}^0=0$.

Over $J$ iterations, HMP-GAE aims to minimize
\begin{align}
\mathcal J^r(\Theta) =\mathcal L_{\mathrm{GAE}}^r +\lambda_{\mathrm{hist}}\mathcal L_{\mathrm{hist}}^r +\lambda_{\mathrm{reg}}\|\Theta\|_2^2,
\label{eq_defense_training}
\end{align}
where $\lambda_{\mathrm{hist}},\lambda_{\mathrm{reg}}\geq0$. The graph $H^r$ and target $\mathbf A^r$ remain fixed while features and embeddings are updated as $\Theta$. 

\subsubsection{Resilience Scoring and Aggregation}

The per-agent structural errors present
\begin{align}
e_{H,k}^r&=\frac{1}{K}\sum_{m=1}^{K} \operatorname{BCE}(H_{km}^r,\widehat H_{km}^r), \nonumber\\
e_{A,k}^r&=\frac{1}{K-1}\sum_{i\neq k} \operatorname{BCE}(\mathbf A_{ki}^r,\widehat{\mathbf A}_{ki}^r).
\label{eq_agent_reconstruction}
\end{align}
Non-negative scores can be normalized by their round-wise median as
\begin{align}
\mathcal N_r(\{a_i\}_{i=1}^{K})_k
=\frac{a_k}{\epsilon+\operatorname{median}_{i=1,\ldots,K}a_i}.
\label{eq_score_normalization}
\end{align}
A score $\zeta_k^r$ can be obtained by
\begin{align}
\zeta_k^r &=\beta_H\mathcal N_r(\{e_{H,i}^r\}_{i=1}^{K})_k +\beta_A\mathcal N_r(\{e_{A,i}^r\}_{i=1}^{K})_k \nonumber\\
&\quad+\beta_{\mathrm{hist}} \mathcal N_r(\{\chi_i^r\}_{i=1}^{K})_k,
\label{eq_anomaly_score}
\end{align}
where $\beta_H,\beta_A,\beta_{\mathrm{hist}}\geq0$ are the non-negative hyperparameters that control each component's contribution. Moreover, the resilience score is given as
\begin{align}
\mathcal Z_k^r=\exp(-\zeta_k^r/\tau),\qquad \tau>0.
\label{eq_trustscore}
\end{align}
As shown in \eqref{eq_global_objective}, larger inconsistencies reduce the agent's contribution through
\begin{align}
\tilde p_k^r=\frac{p_k\mathcal Z_k^r} {\sum_{i=1}^{K}p_i\mathcal Z_i^r}.
\label{eq_defense_weights}
\end{align} 
The server then applies
\begin{align}
\omega^{r+1}=\sum_{k=1}^{K}\tilde p_k^r\widehat\omega\zeta_k^r,
\label{eq_defense_aggregation}
\end{align}
Reconstructed weight changes can be used for assessment, while aggregated factors are used for aggregation. Thus, the server can store $h_k^r=z_k^r$, $\mathcal Z_k^r$, and $\mathcal C^r$, and broadcast $\omega^{r+1}$.

\begin{algorithm}[t]
\caption{HMP-GAE for Byzantine-Resilient FedLLMs}
\label{alg_hmpgae}
\begin{algorithmic}[1]
\INPUT $\omega^r$, $\{\widehat\omega\zeta_k^r\}_{k=1}^{K}$, $\{p_k\}_{k=1}^{K}$, $\{h_k^{r-1}\}_{k=1}^{K}$, and $\Theta$.
\FOR{$k=1,\ldots,K$}
\STATE Reconstruct $U_{k,\ell}^r$ and form $u_k^r$ based on \eqref{eq_reconstructed_change}-\eqref{eq_update_descriptor}.
\STATE \eqref{eq_nodefeature} $\to$ initial node features.
\ENDFOR
\STATE \eqref{eq_incidence}, \eqref{eq_degree}, and \eqref{eq_adjacency_target} $\to$ construct $H^r$, the degree matrices, and $\mathbf A^r$.
\FOR{$t=1,\ldots,J$}
\STATE Obtain node features and apply the $L$ HMP layers.
\STATE Decode $\widehat{\mathbf A}^r$ and $\widehat H^r$.
\IF{$r>0$}
\STATE Re-encode $\mathcal C^{r-1}$ and \eqref{eq_historical_reference}--\eqref{eq_historical_loss} $\to$ $\mathcal L_{\mathrm{hist}}^r$.
\ELSE
\STATE Set $\chi_k^r=0$ and $\mathcal L_{\mathrm{hist}}^r=0$.
\ENDIF
\STATE \eqref{eq_defense_training} $\to$ minimize $\mathcal J^r(\Theta)$.
\ENDFOR
\STATE Update final embeddings, reconstructions, and historical discrepancies with the updated $\Theta$.
\STATE \eqref{eq_agent_reconstruction} $\to$ per-agent errors.
\STATE Obtain $\zeta_k^r$, $\mathcal Z_k^r$, and $\tilde p_k^r$ based on~\eqref{eq_anomaly_score}--\eqref{eq_defense_weights}.
\STATE \eqref{eq_defense_aggregation} $\to$ aggregate $\omega^{r+1}$.
\STATE Keep $\mathcal C^r$, $h_k^r=z_k^r$, and $\mathcal Z_k^r$.
\STATE Broadcast $\omega^{r+1}$.
\end{algorithmic}
\end{algorithm}

\subsection{Complexity Analysis}

Let $\Phi(H)$ denote the number of non-zero incidence entries, and $\Phi(H)=K(s+1)$. Each HMP layer costs $\mathcal O(Kd^2+\Phi(H)d)$, while dense incidence and adjacency decoding each cost $\mathcal O(K^2d)$. Given an input projection from $D+d$ to $d$, historical re-encoding and final scoring pass during $J$ steps cost
\begin{align}
\mathcal O\big((J\!+\!1)[KDd\!+\!L(Kd^2\!\!+\!K(s\!+\!1)d)\!+\!K^2d]\big).
\end{align}
The additional $Kd^2$ feature-processing cost is absorbed by the encoder term for $L\geq1$. 

Explicit LoRA product reconstruction costs $\mathcal O(K\sum_{\ell\in\mathcal S}a_\ell b_\ell q_\ell)$. The neighbor search with linear-time selection and adjacency construction can cost $\mathcal O(K^2d+K(s+1)^2)$, and aggregation
costs $\mathcal O(KP)$, where $P$ contains the total number of scalar entries of $\omega$ in~\eqref{eq_omega}. The total per-round server cost is
\begin{align}
\mathcal O\Big( &K\sum_{\ell\in\mathcal S}a_\ell b_\ell q_\ell +KP+K^2d+K(s+1)^2 +(J+1)\nonumber\\
&\times [KDd+L(Kd^2+K(s+1)d)+K^2d]\Big).
\end{align}
Given $s$ and $d$, the sparse HMP propagation scales linearly with~$K$. 

\section{Performance Evaluation}
\label{sec_performance}
This section presents the implementation of HMP-GAE in FedLLMs under Byzantine attacks. The source code for HMP-GAE is available at: \href{https://github.com/GuangLun2000/HMP-GAE}{https://github.com/GuangLun2000/HMP-GAE}.

\subsection{Experimental Implementation}
The experimental evaluation is conducted on a Linux workstation equipped with an NVIDIA RTX PRO 6000 Blackwell GPU with 96\,GB of memory. The framework is implemented in Python 3.12 with PyTorch 2.10. Two representative text classification datasets \cite{zhang2015character} are adopted to evaluate the proposed HMP-GAE framework:
\begin{itemize}
    \item The first is the \textit{AG News dataset}, which consists of four topical categories, namely World, Sports, Business, and Sci/Tech, containing 120,000 training samples and 7,600 test samples. 
    \item The second one is the \textit{Yahoo! Answers dataset}, a large-scale corpus for topic classification that includes 10 categories, with approximately 1.4 million training instances and 60,000 test samples.
\end{itemize}
These datasets differ in label granularity and corpus scale, helping assess the Byzantine-resilient performance under distinct semantic distributions and learning conditions. 

Two pretrained backbones are deployed at the agents to help examine the effect of model capacity on Byzantine resilience.
\begin{itemize}
    \item Qwen-2.5 \cite{hui2024qwen2} is a model with approximately 0.5 billion parameters, pretrained on 18 trillion tokens with broad multilingual coverage and a context length of up to 32K tokens.
    \item Llama-3.2 \cite{grattafiori2024llama} is a model with around 1.2 billion parameters, pretrained on up to 9 trillion tokens from multilingual online data and enhanced through knowledge distillation from larger Llama models, with a context length of up to 128K tokens.
\end{itemize}
To enable efficient adaptation of these backbones in the federated setting, we employ LoRA, where trainable low-rank matrices are injected into the attention and feedforward projections while the pretrained weights remain frozen. Consequently, each agent optimizes and uploads only the flattened low-rank update, which constitutes both the communication payload and the input to the proposed resilience mechanism at the server.

\begin{table*}[t]
\centering
\small
\renewcommand{\arraystretch}{1.42}
\setlength{\tabcolsep}{6pt}
\caption{\small Global model performance of the proposed HMP-GAE, compared to Multi-Krum, FoolsGold, and FLTrust methods. Accuracy and CSE\textsubscript{mean} are averaged over 50 rounds, while CSE\textsubscript{final} and PPL are evaluated based on the final global model.}
\label{tab:noniid_defense_results}
\begin{tabular}{l l l c c c c | c c c c}
\toprule
\multirow{2}{*}{\textbf{Dataset}} &
\multirow{2}{*}{\textbf{Ref.}} &
\multirow{2}{*}{\textbf{Method}} &
\multicolumn{4}{c}{\textbf{Qwen-2.5} \cite{hui2024qwen2}} &
\multicolumn{4}{c}{\textbf{Llama-3.2} \cite{grattafiori2024llama}} \\
\cmidrule(lr){4-7}
\cmidrule(lr){8-11}
& & &
\textbf{Accuracy} & \textbf{CSE\textsubscript{mean}} & \textbf{CSE\textsubscript{final}} & \textbf{PPL} &
\textbf{Accuracy} & \textbf{CSE\textsubscript{mean}} & \textbf{CSE\textsubscript{final}} & \textbf{PPL} \\
\midrule

\multirow{4}{*}{\textbf{\textit{AG News}}}
& \cite{blanchardByzantineTolerantMachineLearning2017a}
& Multi-Krum
& 85.68 & 0.2830 & 0.0870 & 83.15
& 89.16 & 0.1525 & 0.0539 & 69.16 \\

& \cite{fungMitigatingSybilsFederated2020a}
& FoolsGold
& 88.00 & 0.1608 & 0.0397 & 78.81
& 89.63 & 0.1639 & 0.0290 & 60.04 \\

& \cite{caoFLTrustByzantinerobustFederated2022a}
& FLTrust
& 87.47 & 0.2859 & 0.1870 & 76.08
& 90.23 & 0.2379 & 0.1161 & 53.68 \\

& \textbf{Ours}
& \textbf{HMP-GAE}
& \textbf{88.95} & \textbf{0.1267} & \textbf{0.0395} & \textbf{69.49}
& \textbf{91.44} & \textbf{0.1067} & \textbf{0.0258} & \textbf{46.52} \\

\midrule

\multirow{4}{*}{\shortstack[l]{\textbf{\textit{Yahoo!}}\\\textbf{\textit{Answers}}}}
& \cite{blanchardByzantineTolerantMachineLearning2017a}
& Multi-Krum
& 54.13 & 1.0865 & 0.6247 & 1257.59
& 62.52 & 0.9452 & 0.4692 & 689.66 \\

& \cite{fungMitigatingSybilsFederated2020a}
& FoolsGold
& 60.40 & 0.6978 & 0.3040 & 1549.30
& 65.51 & 0.7357 & 0.4097 & 740.85 \\

& \cite{caoFLTrustByzantinerobustFederated2022a}
& FLTrust
& 56.54 & 1.2385 & 1.1170 & 1128.07
& 64.64 & 1.0694 & 0.8275 & 553.21 \\

& \textbf{Ours}
& \textbf{HMP-GAE}
& \textbf{61.12} & \textbf{0.6378} & \textbf{0.2380} & \textbf{1061.74}
& \textbf{66.54} & \textbf{0.5095} & \textbf{0.2291} & \textbf{525.65} \\

\bottomrule
\end{tabular}
\vspace{10pt}
\end{table*}

\subsection{Performance Metrics}
HMP-GAE is evaluated using three complementary performance metrics, namely prediction accuracy, classification semantic entropy (CSE), and perplexity (PPL), to comprehensively assess its semantic resilience, language modeling capability, and task-specific prediction in FedLLMs.
\begin{itemize}
    \item Prediction accuracy is adopted to evaluate how precisely a FedLLM can predict the target outputs, i.e., class labels, on unseen data after federated aggregation. It is computed as the ratio of correctly estimated samples to the total number of evaluated samples. A higher estimation accuracy indicates that the aggregated FedLLM can correctly preserve task-specific knowledge learned from distributed agents and produce more reliable predictions under data heterogeneity and potentially malicious updates.
    \item CSE is employed to quantify the semantic uncertainty of FedLLMs, which evaluates whether the model assigns predictive mass consistently to a single meaning. It is the mean Shannon entropy of the predictive class distribution over the evaluation set, where the class labels serve as semantic equivalence classes and provide a generation-free surrogate of semantic entropy. Lower CSE indicates that the FedLLM produces semantically stable and confident predictions; higher CSE suggests greater semantic dispersion, uncertainty, or potential Byzantine attacks.
    \item PPL measures how well the aggregated global model predicts a sequence of tokens. A lower perplexity indicates that the FedLLM assigns higher probability to the correct token sequence, suggesting stronger language understanding, generalization, and more effective aggregation across distributed agents.
\end{itemize}

\subsection{Resilience Performance}

The proposed HMP-GAE is compared with three representative robust-aggregation defenses under an identical federated configuration, where the aggregation rule at the server is the only variable.
\begin{itemize}
    \item Multi-Krum \cite{blanchardByzantineTolerantMachineLearning2017a} scores each update by the sum of squared Euclidean distances to its nearest neighbors, and averages the several updates with the lowest scores, thereby retaining only the contributions that reside in the densest region of the parameter space.
    \item FoolsGold \cite{fungMitigatingSybilsFederated2020a} identifies colluding agents by the pairwise cosine similarity of their historical updates, and attenuates the learning rate of the agents whose update directions are persistently aligned with each other.
    \item FLTrust \cite{caoFLTrustByzantinerobustFederated2022a} maintains a clean root dataset at the server to compute a reference update, and weighs each agent by the cosine similarity between its update and this reference after normalizing the update magnitudes.
\end{itemize}

\subsubsection{Performance on The Two Datasets}
Table~\ref{tab:noniid_defense_results} compares HMP-GAE with Multi-Krum, FoolsGold, and FLTrust on \textit{AG News} and \textit{Yahoo! Answers}. HMP-GAE achieves the highest detection accuracy across both datasets and model backbones. On \textit{AG News}, HMP-GAE achieves an accuracy of 88.95\% with Qwen-2.5, outperforming Multi-Krum, FoolsGold, and FLTrust by 3.27\%, 0.95\%, and 1.48\%, respectively. With Llama-3.2, HMP-GAE achieves 91.44\%, improving over the same baselines by 2.28\%, 1.81\%, and 1.21\%. On \textit{Yahoo! Answers}, HMP-GAE achieves 61.12\% with Qwen-2.5 and 66.54\% with Llama-3.2, corresponding to gains of 6.99\%, 0.72\%, and 4.58\%, and 4.02\%, 1.03\%, and 1.90\%, respectively. These gains confirm that our hypergraph message-passing structure sufficiently captures higher-order relationships among multiple connected entities, providing richer contextual evidence for distinguishing corrupted from non-corrupted outputs. The graph autoencoder's reconstruction objective can further encourage representations that preserve shared relational structure, helping separate consistent patterns from anomalous ones and thereby supporting more accurate detection.

\begin{figure*}[t]
    \centering
    \includegraphics[width=2.036 \columnwidth]{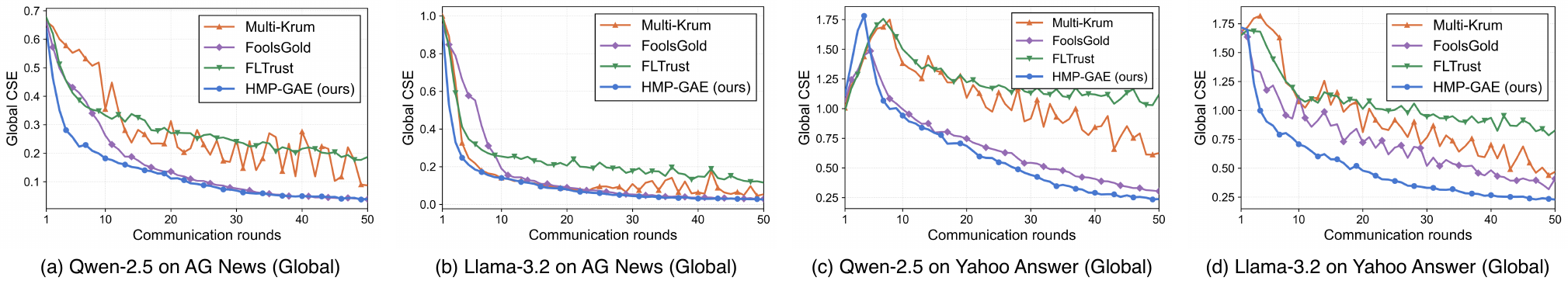}
    \caption{\small Global model CSE of four defenses under the Byzantine attack across 50 communication rounds.}
    \label{fig:cse-global}
\end{figure*}

HMP-GAE achieves the lowest CSE\textsubscript{mean} and CSE\textsubscript{final} in every evaluated setting. On \textit{AG News}, its CSE\textsubscript{mean}/CSE\textsubscript{final} values are 0.1267/0.0395 with Qwen-2.5 and 0.1067/0.0258 with Llama-3.2, representing reductions of 21.21\%/0.50\% and 30.03\%/11.03\% relative to the strongest baseline for each metric. On \textit{Yahoo! Answers}, the corresponding values are 0.6378/0.2380 and 0.5095/0.2291, yielding reductions of 8.60\%/21.71\% and 30.75\%/44.08\%, respectively. Fig.~\ref{fig:cse-global} further shows the global model CSE over 50 communication rounds, where a lower CSE\textsubscript{mean} indicates reduced semantic uncertainty throughout training and a lower CSE\textsubscript{final} shows that this advantage persists in the final global model. Together with the highest accuracy, this indicates that HMP-GAE produces more decisive classifications aligned with the correct labels. Sharper discrimination among the semantic classes is therefore a plausible explanation for this advantage, rather than lower entropy alone being taken as evidence of correctness.

Hypergraph message passing can contribute to this reduction by combining evidence across jointly connected entities, allowing group-level context to resolve ambiguities that individual representations may retain. The autoencoder's reconstruction objective complements this process by building structurally consistent latent representations, providing a plausible basis for the lower semantic uncertainty observed alongside improved accuracy.

In addition, HMP-GAE achieves the lowest final-model PPL across all four settings. On \textit{AG News}, its PPL is 69.49 with Qwen-2.5 and 46.52 with Llama-3.2, representing reductions of 16.43\%, 11.83\%, and 8.66\%, and 32.74\%, 22.52\%, and 13.34\% relative to Multi-Krum, FoolsGold, and FLTrust, respectively. On \textit{Yahoo! Answers}, HMP-GAE reduces PPL to 1061.74 with Qwen-2.5 and 525.65 with Llama-3.2, improving over the same baselines by 15.57\%, 31.47\%, and 5.88\%, and 23.78\%, 29.05\%, and 4.98\%, respectively. As lower PPL indicates that the final global model assigns higher likelihood to the evaluation text, the performance enhancement of PPL is consistent with HMP-GAE retaining useful LLMs information during federated training. Moreover, the graph autoencoder's reconstruction reinforces the retention of these shared patterns in the learned representations, which supports the final global model's language-modeling performance and lower PPL.

\begin{figure*}[t]
    \centering
    \includegraphics[width=2.036 \columnwidth]{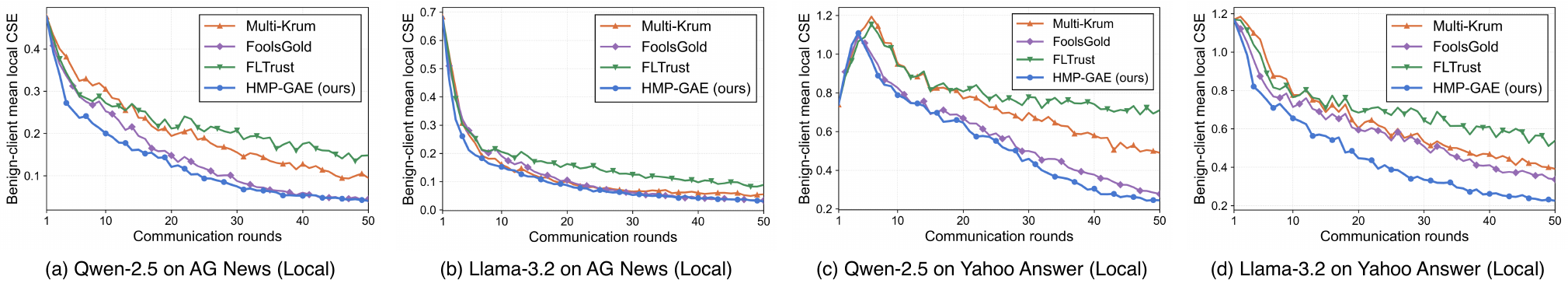}
    \caption{\small Local CSE averaged over all benign agents under four defenses against the Byzantine attack across 50 communication rounds.}
    \label{fig:cse-local}
\end{figure*}

\begin{table}
\centering
\caption{\small Local accuracy (Acc.) and local CSE averaged over all benign local agents under the Byzantine attack.}
\renewcommand{\arraystretch}{1.36}
\setlength{\tabcolsep}{5.4pt}
\label{tab:local_benign_agent_metrics}
\begin{tabular}{l l c c | c c}
\toprule
\multirow{2}{*}{\textbf{Dataset}} &
\multirow{2}{*}{\textbf{Method}} &
\multicolumn{2}{c}{\textbf{Qwen-2.5} \cite{hui2024qwen2}} &
\multicolumn{2}{c}{\textbf{Llama-3.2} \cite{grattafiori2024llama}} \\
\cmidrule(lr){3-4}\cmidrule(lr){5-6}
& & \textbf{Acc.} & \textbf{CSE} & \textbf{Acc.} & \textbf{CSE} \\
\midrule

\multirow{4}{*}{\textbf{\textit{AG News}}}
& Multi-Krum~\cite{blanchardByzantineTolerantMachineLearning2017a}
& 83.69 & 0.2052 & 87.10 & 0.1276 \\

& FoolsGold~\cite{fungMitigatingSybilsFederated2020a}
& 84.76 & 0.1495 & 87.04 & 0.1256 \\

& FLTrust~\cite{caoFLTrustByzantinerobustFederated2022a}
& 83.74 & 0.2259 & 86.76 & 0.1704 \\

& \textbf{HMP-GAE (Ours)}
& \textbf{85.00} & \textbf{0.1257} & \textbf{87.69} & \textbf{0.1085} \\

\midrule

\multirow{4}{*}{\shortstack[l]{\textbf{\textit{Yahoo!}}\\\textbf{\textit{Answers}}}}
& Multi-Krum~\cite{blanchardByzantineTolerantMachineLearning2017a}
& 46.20 & 0.7477 & 52.36 & 0.6443 \\

& FoolsGold~\cite{fungMitigatingSybilsFederated2020a}
& 49.63 & 0.5930 & 54.26 & 0.5910 \\

& FLTrust~\cite{caoFLTrustByzantinerobustFederated2022a}
& 45.98 & 0.8202 & 52.87 & 0.7153 \\

& \textbf{HMP-GAE (Ours)}
& \textbf{50.52} & \textbf{0.5531} & \textbf{55.03} & \textbf{0.4613} \\

\bottomrule
\end{tabular}
\end{table}

\begin{figure*}[t]
    \centering
    \includegraphics[width=2.06\columnwidth]{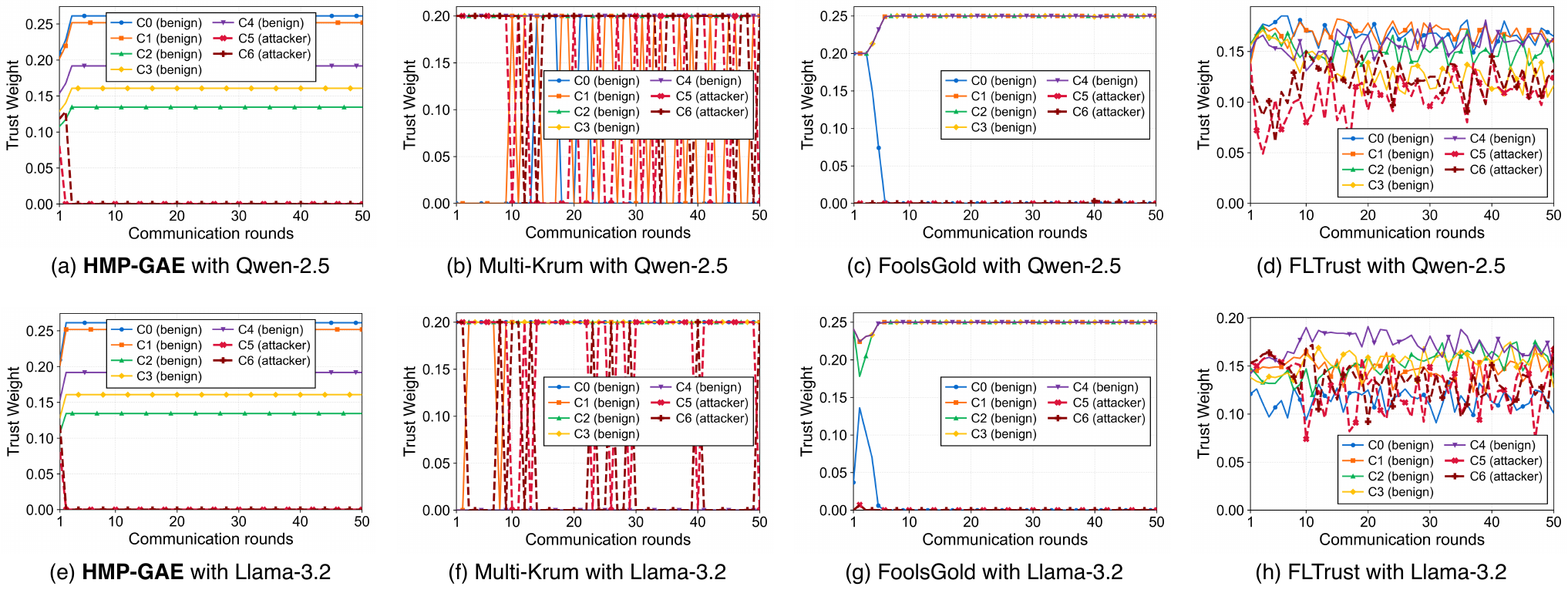}
    \caption{\small Per-user trust weight of four different defenses over 50 communication rounds under the Byzantine attack, with the Qwen-2.5 and Llama-3.2 backbones on \textit{AG News dataset}.}
    \label{fig:trust-weight}
\end{figure*}

Table~\ref{tab:local_benign_agent_metrics} presents that HMP-GAE achieves the highest average local accuracy and the lowest average local CSE among benign agents under the Byzantine attack, where Fig.~\ref{fig:cse-local} shows the variation of local models' CSE over 50 rounds. On \textit{AG News}, the local accuracies of HMP-GAE are 85.00\% and 87.69\% with Qwen-2.5 and Llama-3.2, respectively, surpassing the strongest accuracy baseline in each setting by 0.24\% and 0.59\%. The corresponding local CSE values decrease to 0.1257 and 0.1085, representing reductions of 15.92\% and 13.61\% relative to the lowest-CSE baseline. On \textit{Yahoo! Answers}, HMP-GAE has local accuracies of 50.52\% and 55.03\%, exceeding the strongest baselines by 0.89\% and 0.77\%, while reducing local CSE to 0.5531 and 0.4613, corresponding to reductions of 6.73\% and 21.95\%, respectively. This indicates that HMP-GAE's benefits extend to benign local agents, improving their average prediction accuracy while reducing semantic uncertainty under attack.

\subsection{Per-user Trust Weight}
Fig.~\ref{fig:trust-weight} presents the trust weights assigned to five benign users (C0 to C4) and two malicious users (C5 and C6) on \textit{AG News}. With Qwen-2.5 and Llama-3.2, HMP-GAE reduces the two malicious weights to approximately zero and maintains this suppression throughout the remaining rounds, while assigning stable, positive weights to all five benign users. This clear separation demonstrates the persistent identification of malicious users without excluding benign participants. 

Multi-Krum, however, intermittently assigns malicious users weights of approximately 0.20, equal to those of selected benign users, indicating that malicious updates remain eligible for aggregation in some rounds. FLTrust also fails to separate the two groups: malicious users retain substantial, fluctuating weights that overlap with those of benign ones, particularly with Llama-3.2. Although FoolsGold suppresses two malicious users, it also reduces the benign user C0's weight to approximately zero for most rounds under both model backbones, revealing a persistent false alarm that excludes a legitimate participant's contribution. This confirms HMP-GAE's ability to jointly suppress malicious contributions and retain benign participation, where hypergraph message passing successfully captures higher-order relationships among users, providing collective evidence for distinguishing malicious behavior from benign heterogeneity. The graph autoencoder's reconstruction objective further encourages representations that preserve consistent relational patterns, supporting the sustained separation observed in the trust weights.

\begin{figure*}[t]
    \centering
    \includegraphics[width=1.7 \columnwidth]{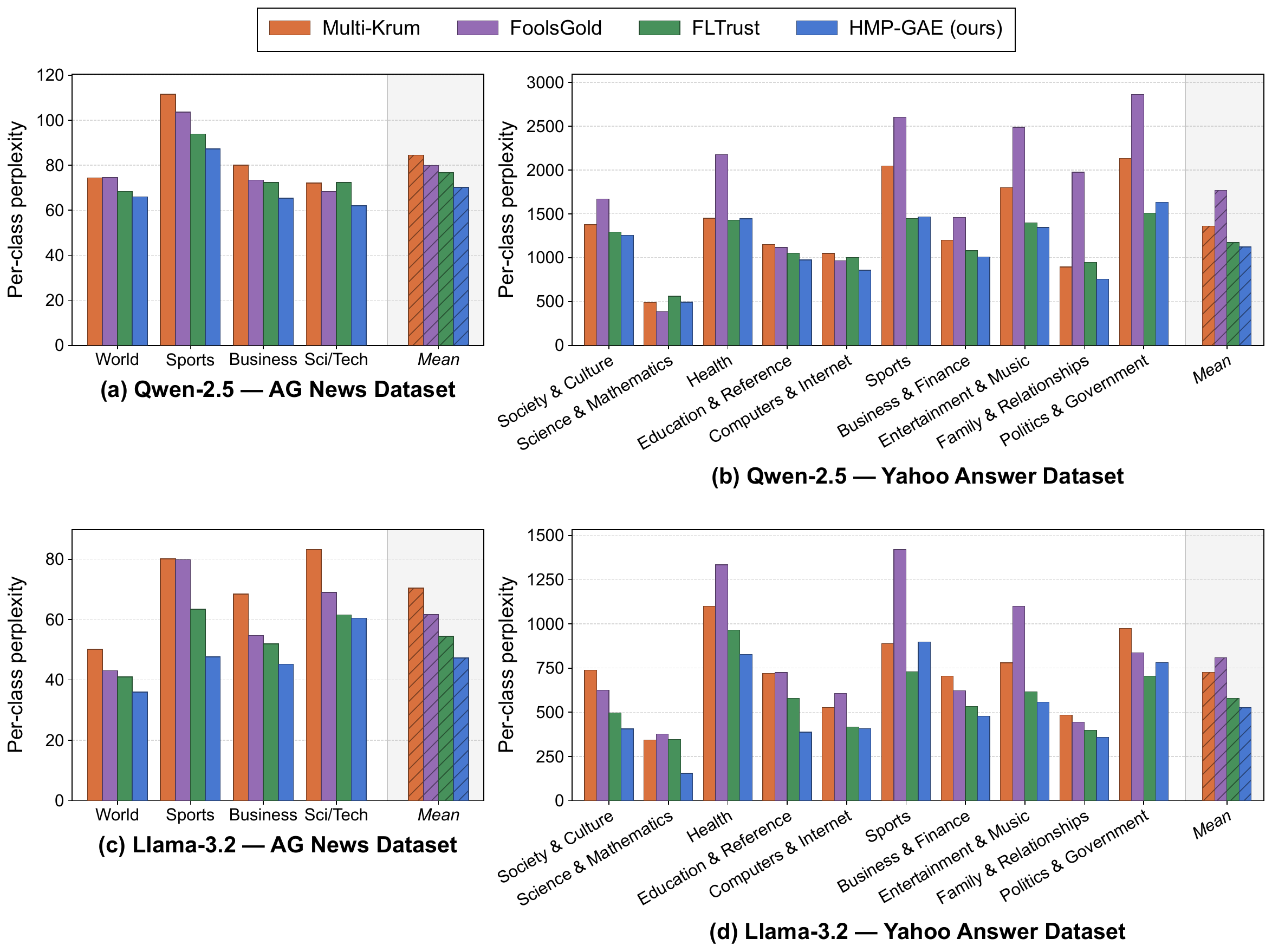}
    \caption{\small Per-class PPL of the final global model under four different defenses against the Byzantine attack, with the Qwen-2.5 and Llama-3.2 backbones on \textit{AG News dataset} and \textit{Yahoo! Answers dataset}.}
    \label{fig:ppl-4four}
\end{figure*}

\begin{table*}[t]
\centering
\caption{\small Ablation study of HMP-GAE under the Byzantine attack, comparing the full HMP-GAE framework with a GAE-based defense without HMP and with HMP-GAE without hyperedge reconstruction. Accuracy and CSE are averaged over 50 rounds, and PPL is evaluated on the final global model.}
\label{tab:ablation}

\renewcommand{\arraystretch}{1.16}
\setlength{\tabcolsep}{4pt}
\small

\begin{tabularx}{\textwidth}{
    @{}
    >{\raggedright\arraybackslash}p{0.11\textwidth}
    >{\raggedright\arraybackslash}p{0.14\textwidth}
    >{\raggedright\arraybackslash}X
    >{\centering\arraybackslash}p{0.13\textwidth}
    >{\centering\arraybackslash}p{0.08\textwidth}
    >{\centering\arraybackslash}p{0.08\textwidth}
    @{}
}
\toprule
\textbf{Model}
& \textbf{Dataset}
& \textbf{Method}
& \textbf{Accuracy (\%) $\uparrow$}
& \textbf{CSE $\downarrow$}
& \textbf{PPL $\downarrow$} \\
\midrule

\multirow{6}{*}{Qwen-2.5}
& \multirow{3}{*}{\textit{AG News}}
& \textit{HMP-GAE}
& \textbf{88.95}
& \textbf{0.1267}
& \textbf{69.49} \\

&
& \textit{GAE-based defense (w/o HMP)}
& 88.58
& 0.1746
& 76.21 \\

&
& \textit{HMP-GAE w/o Hyperedge Reconstruction}
& 88.71
& 0.1476
& 72.75 \\

\cmidrule(lr){2-6}

&
\multirow{3}{*}{\textit{Yahoo! Answers}}
& \textit{HMP-GAE}
& \textbf{61.12}
& \textbf{0.6378}
& \textbf{1061.74} \\

&
& \textit{GAE-based defense (w/o HMP)}
& 59.73
& 1.1034
& 1220.49 \\

&
& \textit{HMP-GAE w/o Hyperedge Reconstruction}
& 60.01
& 0.7449
& 1130.91 \\

\midrule

\multirow{6}{*}{Llama-3.2}
& \multirow{3}{*}{\textit{AG News}}
& \textit{HMP-GAE}
& \textbf{91.44}
& \textbf{0.1067}
& \textbf{46.52} \\

&
& \textit{GAE-based defense (w/o HMP)}
& 91.15
& 0.2344
& 51.73 \\

&
& \textit{HMP-GAE w/o Hyperedge Reconstruction}
& 91.23
& 0.1184
& 49.01 \\

\cmidrule(lr){2-6}

&
\multirow{3}{*}{\textit{Yahoo! Answers}}
& \textit{HMP-GAE}
& \textbf{66.54}
& \textbf{0.5095}
& \textbf{525.65} \\

&
& \textit{GAE-based defense (w/o HMP)}
& 66.28
& 0.8317
& 814.87 \\

&
& \textit{HMP-GAE w/o Hyperedge Reconstruction}
& 66.31
& 0.5363
& 620.23 \\

\bottomrule
\end{tabularx}
\end{table*}

\subsection{Per-class PPL}

In Fig.~\ref{fig:ppl-4four}, the final global model is evaluated across four classes in \textit{AG News} and ten classes in \textit{Yahoo! Answers} using Qwen-2.5 and Llama-3.2. Specifically, HMP-GAE achieves the lowest PPL in all four \textit{AG News} classes with both model backbones, demonstrating improvements across the full range of topics. On \textit{Yahoo! Answers}, HMP-GAE obtains the lowest PPL in most classes and the lowest mean per-class PPL with both backbones, although several class-specific exceptions remain. With Qwen-2.5, FoolsGold achieves lower PPL on Science \& Mathematics, while FLTrust performs slightly better on Health and Sports and more clearly better on Politics \& Government. 

With Llama-3.2, FLTrust achieves lower PPL on Sports and Politics \& Government while Multi-Krum also slightly outperforms HMP-GAE on Sports. These exceptions may reflect differences in class representation across users and in the topic-specific capabilities of the two model backbones. In particular, assigning trust at the user level can reduce the contribution of an untrustworthy user who nevertheless supplies some useful information for a particular class, creating a trade-off between suppressing malicious contributions and retaining class-specific knowledge. Hypergraph message passing and autoencoder reconstruction can favor patterns supported by consistent relationships among users, but such patterns may represent less widely shared topics less effectively. Therefore, the lowest PPL in every individual class is not guaranteed, even when the method improves overall performance. HMP-GAE's lowest mean per-class PPL across all four settings supports its aggregate benefit despite these localized trade-offs.

\subsection{Ablation Study}

Table~\ref{tab:ablation} evaluates the contributions of HMP and hyperedge reconstruction. Specifically, the proposed HMP-GAE achieves the highest accuracy and lowest CSE and PPL across Qwen-2.5 and Llama-3.2 models, as well as the two datasets. Compared with the GAE-based defense, HMP-GAE improves accuracy by $0.26\%$-$1.39\%$, while reducing CSE by $27.43\%$-$54.48\%$, and lowering PPL by $8.82\%$-$35.49\%$. In particular, for Qwen-2.5 on \textit{Yahoo! Answers}, accuracy increases from $59.73\%$ to $61.12\%$. For Llama-3.2, the largest CSE reduction occurs on \textit{AG News}, while the largest PPL reduction occurs on \textit{Yahoo! Answers}. This performance enhancement further confirms the importance of HMP for capturing relationships among multiple (benign and malicious) agents beyond pairwise graph interactions. Such relational information helps assess Byzantine-inducing model updates that may appear consistent individually but deviate from broader group patterns.

Furthermore, compared with HMP-GAE without Hyperedge Reconstruction, the proposed HMP-GAE defense improves accuracy by $0.21\%$-$1.11\%$, reduces CSE by $5.00\%$-$14.38\%$, and lowers PPL by $4.48\%$-$15.25\%$. The most significant accuracy gain and CSE reduction are observed for Qwen-2.5 on \textit{Yahoo! Answers}. For Llama-3.2 on the same dataset, hyperedge reconstruction reduces PPL from $620.23$ to $525.65$. This is because hyperedge reconstruction in HMP-GAE can flag model updates associated with unusually high classification uncertainty before relational assessment. Reducing the influence further limits the propagation of unreliable information during HMP.

\section{Conclusions}
\label{sec_cond}
In this paper, we proposed HMP-GAE, a resilience framework that mitigates Byzantine-inducing updates in FedLLMs through self-healing federated aggregation. We formulated a defensive optimization problem for benign users to maximize global task accuracy while filtering out malicious, semantically manipulated contributions during aggregation. HMP-GAE represents each agent's LoRA-fine-tuned model update as a hypergraph node, with hyperedges capturing higher-order semantic and contextual relationships among multiple updates beyond pairwise correlations. The server iteratively passes messages across these hyperedges to refine benign-update embeddings, enabling collective reasoning to identify and suppress Byzantine-inducing or semantically inconsistent contributions. Experiments on AG News and Yahoo! Answers using Qwen-2.5 and Llama-3.2 demonstrated that HMP-GAE consistently achieved higher accuracy and lower classification semantic entropy and overall perplexity than the evaluated baselines. Trust-weight analysis further showed that HMP-GAE persistently suppressed malicious users' contributions while preserving benign participation, with benefits extending to benign agents' local performance.


\ifCLASSOPTIONcaptionsoff
  \newpage
\fi

\bibliographystyle{IEEEtran}
\bibliography{bibDefALLM}

@article{he2024g,
	author = {He, Xiaoxin and Tian, Yijun and Sun, Yifei and Chawla, Nitesh V and Laurent, Thomas and LeCun, Yann and Bresson, Xavier and Hooi, Bryan},
	journal = {Advances in Neural Information Processing Systems},
	pages = {132876--132907},
	title = {G-retriever: Retrieval-augmented generation for textual graph understanding and question answering},
	volume = {37},
	year = {2024}}

@article{tao2026byzantine,
	author = {Tao, Youming and Zhang, Zuyuan and Wang, Di and Yu, Dongxiao and Cheng, Xiuzhen and Dressler, Falko},
	journal = {IEEE Transactions on Networking},
	publisher = {IEEE},
	title = {Byzantine-Resilient Federated Learning under Heterogeneity and Heavy Tails},
	year = {2026}}

@article{telyatnikov2023hypergraph,
	author = {Telyatnikov, Lev and Bucarelli, Maria Sofia and Bernardez, Guillermo and Zaghen, Olga and Scardapane, Simone and Li{\'o}, Pietro},
	journal = {Transactions on Machine Learning Research},
	title = {Hypergraph neural networks through the lens of message passing: A common perspective to homophily and architecture design},
	year = {2025}}

@inproceedings{zhang2025hypergct,
	author = {Zhang, Xiyu and Ma, Jiayi and Guo, Jianwei and Hu, Wei and Qi, Zhaoshuai and Hui, Fei and Yang, Jiaqi and Zhang, Yanning},
	booktitle = {2025 IEEE/CVF International Conference on Computer Vision (ICCV)},
	organization = {IEEE},
	pages = {24750--24759},
	title = {Hypergct: A dynamic hyper-gnn-learned geometric constraint for 3d registration},
	year = {2025}}

@article{wang2024flora,
	author = {Wang, Ziyao and Shen, Zheyu and He, Yexiao and Sun, Guoheng and Wang, Hongyi and Lyu, Lingjuan and Li, Ang},
	journal = {Advances in Neural Information Processing Systems},
	pages = {22513--22533},
	title = {Flora: Federated fine-tuning large language models with heterogeneous low-rank adaptations},
	volume = {37},
	year = {2024}}

@inproceedings{yan2026fedsrd,
	author = {Yan, Guochen and Xie, Luyuan and Shen, Qingni and Fang, Yuejian and Wu, Zhonghai},
	booktitle = {Proceedings of the ACM Web Conference 2026},
	pages = {5087--5098},
	title = {FedSRD: Sparsify-Reconstruct-Decompose for Communication-Efficient Federated Large Language Models Fine-Tuning},
	year = {2026}}

@article{li2026model,
	author = {Li, Yue and Bhujel, Sudip and Lira, Cameron and Wang, Ning and Xiao, Yang},
	journal = {arXiv preprint arXiv:2608.22552},
	title = {Model-Consistent Byzantine-Resilient Decentralized Federated Learning for Collaborative Missions},
	year = {2026}}

@article{herath2025dsfl,
	author = {Herath, Charuka and Rahulamathavan, Yogachandran and De Silva, Varuna and Lambotharan, Sangarapillai},
	journal = {arXiv preprint arXiv:2509.08449},
	title = {DSFL: A Dual-Server Byzantine-Resilient Federated Learning Framework via Group-Based Secure Aggregation},
	year = {2025}}

@article{yu2025lpp,
	author = {Yu, Jiguo and Zhang, Hongliang and Xia, Qi and Zou, Yifei and Liu, Yangyang},
	journal = {IEEE Transactions on Information Forensics and Security},
	publisher = {IEEE},
	title = {LPP-FL: A Lightweight Privacy-Preserving Federated Learning Against Byzantine Attacks on Non-IID Data},
	year = {2025}}

@article{zhang2026building,
	author = {Zhang, Heyi and He, Xinlei and Wu, Jun and Wang, Qian},
	journal = {IEEE Transactions on Information Forensics and Security},
	publisher = {IEEE},
	title = {Building trust beyond update divergence: dual-refined aggregation for byzantine-robust federated learning},
	year = {2026}}

@article{qin2026efficient,
	author = {Qin, Xian and Yang, Xue and Tang, Xiaohu},
	journal = {IEEE Transactions on Information Forensics and Security},
	publisher = {IEEE},
	title = {Efficient Byzantine-robust privacy-preserving federated learning via dimension compression},
	year = {2026}}

@article{zhu2026bpflh,
	author = {Zhu, Guofu and Shen, Wenting and Liu, Zhiquan and Qin, Jing and Ma, Jixin},
	journal = {IEEE Transactions on Dependable and Secure Computing},
	publisher = {IEEE},
	title = {BPFLH: Byzantine-Robust Privacy-Preserving Federated Learning for Heterogeneous Data},
	year = {2026}}

@article{yazdinejad2024robust,
	author = {Yazdinejad, Abbas and Dehghantanha, Ali and Karimipour, Hadis and Srivastava, Gautam and Parizi, Reza M},
	journal = {IEEE Transactions on Information Forensics and Security},
	pages = {6693--6708},
	publisher = {IEEE},
	title = {A robust privacy-preserving federated learning model against model poisoning attacks},
	volume = {19},
	year = {2024}}

@article{xiao2025flgym,
	author = {Xiao, Ke and Wang, Qiyuan and Anagnostopoulos, Christos},
	journal = {IEEE Transactions on Information Forensics and Security},
	pages = {404--416},
	publisher = {IEEE},
	title = {FLgym: Toward Robust and Byzantine-Resilient Federated Learning},
	volume = {21},
	year = {2025}}

@article{wang2025pdfl,
	author = {Wang, Gang and Wu, Huiwen and Li, Qingming and Liu, Ziyao and Zhao, Jun and Lam, Kwok-Yan and Dong, Qingkuan and Yang, Yao and Zhou, Li},
	journal = {IEEE Transactions on Information Forensics and Security},
	pages = {608--623},
	publisher = {IEEE},
	title = {PDFL: A Privacy-Enhancing and Robust Poisoning Defense Federated Learning Scheme},
	volume = {21},
	year = {2025}}

@article{qiao2025survey,
	author = {Qiao, Yu and Le, Huy and Raha, Avi and Tran, Phuong-Nam and Adhikary, Apurba and Zhang, Mengchun and Nguyen, Loc X and Huh, Eui-Nam and Niyato, Dusit and Hong, Choong Seon},
	journal = {ACM Computing Surveys},
	publisher = {ACM New York, NY},
	title = {A Survey on Foundation Models for Personalized Federated Intelligence},
	year = {2025}}

@article{ren2025advances,
	author = {Ren, Chao and Yu, Han and Peng, Hongyi and Tang, Xiaoli and Zhao, Bo and Yi, Liping and Tan, Alysa Ziying and Gao, Yulan and Li, Anran and Li, Xiaoxiao and others},
	journal = {IEEE Communications Surveys \& Tutorials},
	pages = {2087--2126},
	publisher = {IEEE},
	title = {Advances and open challenges in federated foundation models},
	volume = {28},
	year = {2025}}

@article{xia2026perfect,
	author = {Xia, Yue and Hofmeister, Christoph and Egger, Maximilian and Bitar, Rawad},
	journal = {IEEE Transactions on Information Forensics and Security},
	publisher = {IEEE},
	title = {Perfect Privacy for Discriminator-Based Byzantine-Resilient Federated Learning},
	year = {2026}}

@article{uddin2025systematic,
	author = {Uddin, Md Palash and Xiang, Yong and Hasan, Mahmudul and Bai, Jun and Zhao, Yao and Gao, Longxiang},
	journal = {ACM Computing Surveys},
	number = {10},
	pages = {1--62},
	publisher = {ACM New York, NY},
	title = {A systematic literature review of robust federated learning: Issues, solutions, and future research directions},
	volume = {57},
	year = {2025}}

@article{wang2026byzantine,
	author = {Wang, Jinbo and Wang, Ruijin and Li, Zengpeng and Zhang, Fengli and Li, Jingwei and Li, Xiong},
	journal = {IEEE Transactions on Information Forensics and Security},
	publisher = {IEEE},
	title = {A Byzantine-Robust Secure Federated Learning Scheme in Heterogeneous Data},
	year = {2026}}

@article{song2025byzantine,
	author = {Song, Anxiao and Zhang, Tao and Cheng, Ke and Cao, Yang and Zhu, Xinghui and Shen, Yulong},
	journal = {IEEE Internet of Things Journal},
	number = {14},
	pages = {29073--29088},
	publisher = {IEEE},
	title = {Byzantine-robust federated learning framework via a server-client defense mechanisms},
	volume = {12},
	year = {2025}}

@article{shen2026byzantine,
	author = {Shen, Meng and Peng, Bohan and Zhao, Yi and Qin, Yue and Li, Ming and Li, Qi and Zhu, Liehuang},
	journal = {IEEE Transactions on Information Forensics and Security},
	publisher = {IEEE},
	title = {Byzantine-Robust Asynchronous Federated Learning via Feature Fingerprinting},
	year = {2026}}

@article{tang2025roby,
	author = {Tang, Xiangyun and Li, Minyang and Shen, Meng and Kang, Jiawen and Zhu, Liehuang and Liu, Zhiquan and Yang, Guomin and Niyato, Dusit and Deng, Robert H},
	journal = {IEEE Transactions on Information Forensics and Security},
	publisher = {IEEE},
	title = {Roby: A byzantine-robust and privacy-preserving serverless federated learning framework},
	year = {2025}}

@article{xia2026communication,
	author = {Xia, Yuanjun and Tan, Zhiyuan and Zhang, Pengfei and Yu, Lele and Dong, Shi and Liu, Yining},
	journal = {IEEE Transactions on Mobile Computing},
	publisher = {IEEE},
	title = {Communication-Efficient and Byzantine-Tolerant Federated Learning for Multi-Access Edge Computing},
	year = {2026}}

@article{dong2023privacy,
	author = {Dong, Caiqin and Weng, Jian and Li, Ming and Liu, Jia-Nan and Liu, Zhiquan and Cheng, Yudan and Yu, Shui},
	journal = {IEEE Transactions on Dependable and Secure Computing},
	number = {2},
	pages = {889--904},
	publisher = {IEEE},
	title = {Privacy-preserving and Byzantine-robust federated learning},
	volume = {21},
	year = {2023}}

@article{huang2024federated,
	author = {Huang, Wenke and Ye, Mang and Shi, Zekun and Wan, Guancheng and Li, He and Du, Bo and Yang, Qiang},
	journal = {IEEE Transactions on Pattern Analysis and Machine Intelligence},
	number = {12},
	pages = {9387--9406},
	publisher = {IEEE},
	title = {Federated learning for generalization, robustness, fairness: A survey and benchmark},
	volume = {46},
	year = {2024}}

@article{zhang2025practical,
	author = {Zhang, Baolei and Fang, Minghong and Liu, Zhuqing and Yi, Biao and Zhou, Peizhao and Wang, Yuan and Li, Tong and Liu, Zheli},
	journal = {IEEE Transactions on Information Forensics and Security},
	pages = {61--75},
	publisher = {IEEE},
	title = {Practical framework for privacy-preserving and byzantine-robust federated learning},
	volume = {21},
	year = {2025}}

@article{cai2026graph1,
	author = {Cai, Hanlin and Li, Kai and Wang, Houtianfu and Dong, Haofan and Li, Yichen and Dressler, Falko and Akan, Ozgur B},
	journal = {arXiv preprint arXiv:2605.07961},
	title = {Graph Representation Learning Augmented Model Manipulation on Federated Fine-Tuning of LLMs},
	year = {2026}}

@inproceedings{cai2026graph2,
	author = {Cai, Hanlin and Dong, Haofan and Wang, Houtianfu and Li, Kai and Zou, Sai and Akan, Ozgur B},
	booktitle = {2026 International Wireless Communications and Mobile Computing (IWCMC)},
	organization = {IEEE},
	pages = {956--961},
	title = {Graph representation-based model poisoning on the heterogeneous Internet of Agents},
	year = {2026}}

@article{grattafiori2024llama,
	author = {Grattafiori, Aaron and Dubey, Abhimanyu and Jauhri, Abhinav and Pandey, Abhinav and Kadian, Abhishek and Al-Dahle, Ahmad and Letman, Aiesha and Mathur, Akhil and Schelten, Alan and Vaughan, Alex and others},
	journal = {arXiv preprint arXiv:2407.21783},
	title = {The llama 3 herd of models},
	year = {2024}}

@article{wang2025federated,
	author = {Wang, Zixin and Zhou, Yong and Shi, Yuanming and Letaief, Khaled B},
	journal = {IEEE Transactions on Wireless Communications},
	publisher = {IEEE},
	title = {Federated fine-tuning for pre-trained foundation models over wireless networks},
	year = {2025}}

@article{hu2024federated,
	author = {Hu, Jiahui and Wang, Dan and Wang, Zhibo and Pang, Xiaoyi and Xu, Huiyu and Ren, Ju and Ren, Kui},
	journal = {IEEE Wireless Communications},
	publisher = {IEEE},
	title = {Federated large language model: Solutions, challenges and future directions},
	year = {2024}}

@article{yao2024federated,
	author = {Yao, Yuhang and Zhang, Jianyi and Wu, Junda and Huang, Chengkai and Xia, Yu and Yu, Tong and Zhang, Ruiyi and Kim, Sungchul and Rossi, Ryan and Li, Ang and others},
	journal = {arXiv preprint arXiv:2409.15723},
	title = {Federated large language models: Current progress and future directions},
	year = {2024}}

@article{yang2025llm,
	author = {Yang, Hanqing and Chen, Jingdi and Siew, Marie and Lorido-Botran, Tania and Joe-Wong, Carlee},
	journal = {arXiv preprint arXiv:2502.05453},
	title = {Llm-powered decentralized generative agents with adaptive hierarchical knowledge graph for cooperative planning},
	year = {2025}}

@article{webb2025brain,
	author = {Webb, Taylor and Mondal, Shanka Subhra and Momennejad, Ida},
	journal = {Nature Communications},
	number = {1},
	pages = {8633},
	publisher = {Nature Publishing Group UK London},
	title = {A brain-inspired agentic architecture to improve planning with LLMs},
	volume = {16},
	year = {2025}}

@article{li2025future,
	author = {Li, Haiyuan and Madhukumar, Hari and Methley, Nicholas and Chen, Xuewen and Wu, Yulei and Parra-Ullauri, Juan and Sharma, Vishnu and Lee, Jeongran and Koblitz, Arndt Ryo and Andrews, Matthew and others},
	journal = {IEEE Internet of Things Journal},
	publisher = {IEEE},
	title = {Future Factories with 6G: Agentic AI and Cyber-Physical Digital Twins},
	year = {2025}}

@article{cheng2024towards,
	author = {Cheng, Yujun and Zhang, Weiting and Zhang, Zhewei and Zhang, Chuan and Wang, Shengjin and Mao, Shiwen},
	journal = {IEEE Communications Surveys \& Tutorials},
	publisher = {IEEE},
	title = {Towards federated large language models: Motivations, methods, and future directions},
	year = {2024}}

@article{cai2025graph,
	author = {Cai, Hanlin and Dong, Haofan and Wang, Houtianfu and Li, Kai and Akan, Ozgur B},
	journal = {arXiv preprint arXiv:2507.01694},
	title = {Graph Representation-based Model Poisoning on Federated LLMs in CyberEdge Networks},
	year = {2025}}

@article{li2025towards,
	author = {Li, Kai and Zhang, Zhengyang and Pourkabirian, Azadeh and Ni, Wei and Dressler, Falko and Akan, Ozgur B},
	journal = {arXiv preprint arXiv:2504.01240},
	title = {Towards Resilient Federated Learning in CyberEdge Networks: Recent Advances and Future Trends},
	year = {2025}}

@article{zhang2015character,
	author = {Zhang, Xiang and Zhao, Junbo and LeCun, Yann},
	journal = {Advances in neural information processing systems},
	title = {Character-level convolutional networks for text classification},
	volume = {28},
	year = {2015}}

@article{hui2024qwen2,
	author = {Hui, Binyuan and Yang, Jian and Cui, Zeyu and Yang, Jiaxi and Liu, Dayiheng and Zhang, Lei and Liu, Tianyu and Zhang, Jiajun and Yu, Bowen and Lu, Keming and others},
	journal = {arXiv preprint arXiv:2409.12186},
	title = {Qwen2. 5-coder technical report},
	year = {2024}}

@article{blanchardByzantineTolerantMachineLearning2017a,
	archiveprefix = {arXiv},
	author = {Blanchard, Peva and Mhamdi, El Mahdi El and Guerraoui, Rachid and Stainer, Julien},
	eprint = {1703.02757},
	journal = {arXiv preprint arXiv:1703.02757},
	month = mar,
	number = {arXiv:1703.02757},
	primaryclass = {cs.DC},
	publisher = {arXiv},
	title = {Byzantine-{{Tolerant Machine Learning}}},
	year = 2017}

@article{caoFLTrustByzantinerobustFederated2022a,
	archiveprefix = {arXiv},
	author = {Cao, Xiaoyu and Fang, Minghong and Liu, Jia and Gong, Neil Zhenqiang},
	eprint = {2012.13995},
	journal = {arXiv preprint arXiv:2012.13995},
	month = apr,
	number = {arXiv:2012.13995},
	primaryclass = {cs.CR},
	publisher = {arXiv},
	title = {{{FLTrust}}: {{Byzantine-robust Federated Learning}} via {{Trust Bootstrapping}}},
	year = 2022}

@article{fungMitigatingSybilsFederated2020a,
	archiveprefix = {arXiv},
	author = {Fung, Clement and Yoon, Chris J. M. and Beschastnikh, Ivan},
	eprint = {1808.04866},
	journal = {arXiv preprint arXiv:1808.04866},
	month = jul,
	number = {arXiv:1808.04866},
	primaryclass = {cs.LG},
	publisher = {arXiv},
	title = {Mitigating {{Sybils}} in {{Federated Learning Poisoning}}},
	year = 2020}

@inproceedings{cai2024securing,
  title={Securing billion bluetooth devices leveraging learning-based techniques},
  author={Cai, Hanlin},
  booktitle={Proceedings of the AAAI Conference on Artificial Intelligence},
  volume={38},
  number={21},
  pages={23731--23732},
  year={2024}}

@inproceedings{cai2026poster,
  title={Poster: Graph Learning-based Update Manipulation Attack on Federated Fine-Tuning of LLMs over Wireless Networks},
  author={Cai, Hanlin and Zhan, Junfei and Lin, Zheng and Cai, Xinyi},
  booktitle={Proceedings of the 24th Annual International Conference on Mobile Systems, Applications and Services Companion},
  pages={27--28},
  year={2026}
}

\end{document}